\documentclass[sigconf]{acmart}
\AtBeginDocument{%
  }

\setcopyright{acmlicensed}
\renewcommand\footnotetextcopyrightpermission[1]{}
\copyrightyear{2025}
\acmYear{2025}
\acmDOI{XXXXXXX.XXXXXXX}
\usepackage{multirow}
\usepackage{graphicx}
\usepackage{subcaption}
\usepackage{float}
\usepackage{enumitem}
\begin{document}

\title{LLM4CD: Leveraging Large Language Models for Open-World Knowledge Augmented Cognitive Diagnosis}

\author{Weiming Zhang}
\affiliation{%
  \institution{Shanghai Jiao Tong University}
  \city{Shanghai}
  \country{China}
}
\email{WeimingZhang_2020@sjtu.edu.cn}

\author{Lingyue Fu}

\affiliation{%
  \institution{Shanghai Jiao Tong University}
  \city{Shanghai}
  \country{China}
}
\email{fulingyue@sjtu.edu.cn}

\author{Qingyao Li}
\affiliation{%
  \institution{Shanghai Jiao Tong University}
  \city{Shanghai}
  \country{China}
}
\email{ly890306@sjtu.edu.cn}

\author{Kounianhua Du}
\affiliation{%
  \institution{Shanghai Jiao Tong University}
  \city{Shanghai}
  \country{China}
}
\email{kounianhuadu@sjtu.edu.cn}

\author{Jianghao Lin}
\affiliation{%
  \institution{Shanghai Jiao Tong University}
  \city{Shanghai}
  \country{China}
}
\email{chiangel@sjtu.edu.cn}

\author{Jingwei Yu}
\affiliation{%
  \institution{Shanghai Jiao Tong University}
  \city{Shanghai}
  \country{China}
}
\email{yujingwei_2569@sjtu.edu.cn}

\author{Wei Xia}
\affiliation{
\institution{www.imxwell.com}
\city{Shenzhen}
\country{China}
}
\email{xwell.xia@gmail.com}

\author{Weinan Zhang}
\affiliation{%
  \institution{Shanghai Jiao Tong University}
  \city{Shanghai}
  \country{China}
}
\email{wnzhang@sjtu.edu.cn}

\author{Ruiming Tang}
\affiliation{
\institution{Huawei Noah's Ark Lab}
\city{Shenzhen}
\country{China}
}
\email{tangruiming@huawei.com}

\author{Yong Yu}
\affiliation{%
  \institution{Shanghai Jiao Tong University}
  \city{Shanghai}
  \country{China}
}
\email{yyu@apex.sjtu.edu.cn}
\authornote{The corresponding author.}

\renewcommand{\shortauthors}{Zhang et al.}

\begin{abstract}

Cognitive diagnosis (CD) plays a crucial role in intelligent education, evaluating students' comprehension of knowledge concepts based on their test histories. However, current CD methods often model students, exercises, and knowledge concepts solely on their ID relationships, neglecting the abundant semantic relationships present within educational data space. Furthermore, contemporary intelligent tutoring systems (ITS) frequently involve the addition of new students and exercises, creating cold-start scenarios that ID-based methods find challenging to manage effectively. The advent of large language models (LLMs) offers the potential for overcoming this challenge with open-world knowledge. In this paper, we propose \textit{LLM4CD}, which Leverages \underline{L}arge \underline{L}anguage \underline{M}odels \underline{for} open-world knowledge Augmented \underline{C}ognitive \underline{D}iagnosis. Our method utilizes the open-world knowledge of LLMs to construct cognitively expressive textual representations, which are then encoded to introduce rich semantic information into the CD task. Additionally, we propose an innovative bi-level encoder framework that models students' test histories through two levels of encoders: a macro-level cognitive text encoder and a micro-level knowledge state encoder. This approach substitutes traditional ID embeddings with semantic representations, enabling the model to accommodate new students and exercises with open-world knowledge and address the cold-start problem. Extensive experimental results demonstrate that LLM4CD consistently outperforms previous CD models on multiple real-world datasets, validating the effectiveness of leveraging LLMs to introduce rich semantic information into the CD task.

\end{abstract}


\maketitle

\section{INTRODUCTION}
\begin{figure}[t]
  \centering
  \includegraphics[width=\linewidth]{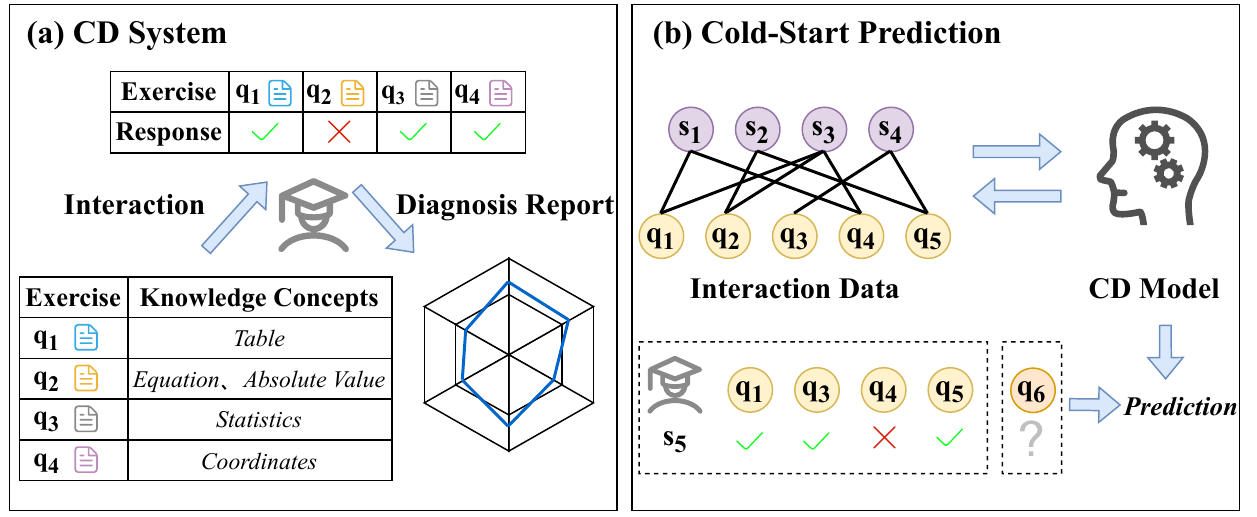} 
  \caption{(a) An illustration of the cognitive diagnosis (CD) task with semantic concepts; (b) An example of the cold-start problem on traditional ID-based CD models.}
  \label{fig:intro}
\end{figure}

Online education platforms, such as Massive Open Online Courses (MOOCs)\footnote{https://www.mooc.org/}, are transforming the landscape of learning, making education more accessible and personalized.  Cognitive Diagnosis (CD) is an essential task in intelligent tutoring systems (ITS)~\cite{liu2021towards, burns2014intelligent},  serving as a crucial component of computerized adaptive testing (CAT)~\cite{wainer2000computerized, ghosh2021bobcat} and resource recommendation~\cite{kuh2011piecing, li2023graph}. CD models estimate students' knowledge states based on their test history and predict their responses to future exercises. As illustrated in Figure~\ref{fig:intro}(a), the CD task involves students responding to some exercises, and the system subsequently inferring their knowledge states to generate a diagnostic report.

CD models have evolved significantly with the development of ITS. In the early stages, CD models primarily relied on manually designed linear functions, as seen in Item Response Theory (IRT)~\cite{embretson2013item} and Multidimensional Item Response Theory (MIRT)~\cite{ackerman2014multidimensional}. These traditional models estimated students' abilities based on their responses to items, assuming a logistic function that related a student's ability to the properties of exercises, such as difficulty and discrimination. With the advent of deep learning~\cite{lecun2015deep}, Neural Cognitive Diagnosis (NCD)~\cite{wang2020neural} emerges, employing neural networks to capture non-linear interactions among students, exercises, and cognitive attributes, thereby enhancing flexibility and diagnostic accuracy. More recently, Relation Map Driven Cognitive Diagnosis (RCD) model~\cite{gao2021rcd} introduces relational graphs to CD, overcoming the challenge of modeling complex educational interactions by utilizing exercise-knowledge correlation and dependency graphs. Despite their effectiveness, current CD models are predominantly ID-based, and overlook the utilization of textual information in educational settings, therefore leading to the following limitations.

Firstly, previous CD models have overlooked the deep semantic relationships present in educational datasets. Concept names tend to contain complex semantic relationships that can convey specific details about exercises and reveal intricate interdependencies among concepts, offering valuable insights into students' knowledge states. However, despite the richness of semantic information embedded in concept names, previous work has largely failed to exploit this potential fully. This limitation is partly due to the limited textual data in educational datasets, often restricted to concept names and a small set of concepts. For instance, as shown in Figure~\ref{fig:intro}(a), the concept name corresponding to $e_1$, "table", might seem ambiguous in general contexts. But within an educational setting, it is contextually linked to mathematical statistics. This connection extends to the "statistics" concept associated with exercise $e_3$. This example highlights the latent potential of concept names in educational datasets and emphasizes the necessity for advanced methods to effectively harness such textual information, thereby improving the accuracy and depth of CD tasks.

% Secondly, existing methods based on ID modeling generally suffer from the cold-start problem~\cite{gao2023leveraging,wei2017collaborative}.  Current CD models that use ID embeddings to represent students and exercises have significant limitations when facing new exercises or students not in the training data. As shown in Figure~\ref{fig:intro}(b), these models struggle with accurate predictions for new entries due to their reliance on pre-existing ID embeddings and limited ability to generalize to unseen data. This dependence on historical IDs results in poor adaptability and accuracy, particularly evident in the cold-start problem, where the introduction of unseen data significantly hinders model performance. To address this issue, CD models must integrate textual information to develop more robust and adaptable representations as new exercises and students are continuously introduced.

Secondly, existing methods based on ID modeling generally suffer from the cold-start problem~\cite{gao2023leveraging,wei2017collaborative}, which arises when new students or exercises, such as $s_1$ and $q_5$ in Figure~\ref{fig:intro}(b), are introduced into ITS. These methods rely on pre-existing ID embeddings (e.g., $q_1$ to $q_4$) to represent students and exercises, making it challenging to generalize to unseen data that lack corresponding ID embeddings in the training set. This inherent limitation often results in degraded performance in cold-start scenarios, where models struggle to provide accurate predictions for new students or exercises. To address this issue, recent research has explored several innovative approaches, such as leveraging transferable knowledge concept graphs~\cite{gao2023leveraging}, utilizing early-bird students for zero-shot diagnosis~\cite{gao2024zero}, and exploring inductive learning frameworks~\cite{liu2024inductive}. However, these methods primarily focus on structured representations like graph-based or ID-driven frameworks, which often neglect the semantic richness and contextual information inherent in educational data. This limits their ability to fully capture students' cognitive states when generalizing to new scenarios. When new exercises are introduced, their associated textual descriptions frequently include rich details such as question types and difficulty levels. These features are highly informative for estimating students' responses and understanding their cognitive states. However, the potential of such rich textual information remains underutilized as input to CD models.

% To address these challenges, it is essential to explore text-based modeling approaches that utilize semantic information for  modeling students' cognition. Such approach can reduce training costs while enhancing the system's flexibility and effectiveness in handling new and diverse educational data, thereby overcoming the limitations of conventional methods. The rapid advancement of large language models (LLMs)~\cite{zhao2023survey} has increasingly highlighted their potential to enhance user modeling with open-world knowledge~\cite{huang2022towards,wei2022chain,lin2023can,tan2023user}, opening new avenues for improving the performance of CD models. 

To address these challenges, it is crucial to incorporate textual information into CD tasks, as it can effectively handle new and diverse educational data by leveraging the inherent semantic richness of educational texts. Such an approach not only enhances the system's flexibility but also reduces training costs by providing a more generalizable framework for modeling students' cognition. While prior works have made attempts to incorporate textual information into CD tasks~\cite{cheng2019dirt,wang2020neural,song2023deep}, these methods are often constrained by their reliance on fixed or limited textual representations, which fail to fully utilize the semantic and structural information embedded in educational datasets.
The rapid advancement of large language models (LLMs)~\cite{zhao2023survey} provides a promising solution, as LLMs can dynamically retrieve and integrate open-world knowledge~\cite{huang2022towards,wei2022chain,lin2023can,tan2023user}, enabling CD models to overcome these limitations. By leveraging LLMs, CD models can better address challenges such as the cold-start problem while achieving more robust and adaptable performance in diverse educational scenarios.

To this end, in this paper, we propose \textit{LLM4CD}, which Leverages \underline{L}arge \underline{L}anguage \underline{M}odels \underline{for} Open-World Knowledge Augmented \underline{C}ognitive \underline{D}iagnosis. LLM4CD effectively addresses the two critical limitations of traditional CD models by leveraging the open-world knowledge capabilities of LLMs. First, it performs deep semantic exploration of concept names, enriching them with cognitive meaning that was previously underutilized. By introducing this deep semantic information into the CD domain, LLM4CD overcomes the limitation of prior models that failed to capture the rich, context-specific meanings inherent in educational data. Additionally, LLM4CD further addresses the second challenge by employing a bi-level encoder framework that models students at both macro and micro levels. This approach replaces traditional ID embeddings with rich textual representations, allowing the model to adapt more effectively to new students and exercises. By leveraging text-based information, LLM4CD overcomes the limitations of ID-based models and is better equipped to handle the cold-start problem, making it highly adaptable to the evolving needs of modern ITS.
% \ljh{how do we mitigate the aforementioned two challenges, this must be stated. saying what the model does is not enough}

In summary, our contributions are listed as follows:
\begin{itemize}[leftmargin=10pt,topsep=5pt]
\item  To the best of our knowledge, we are the first to reveal the critical importance of textual information in CD, which has been overlooked by previous work, and propose integrating LLMs' open-world knowledge into CD tasks to bridge this gap.
% \item \ljh{we propose the first LLM-based framework}
\item We propose \textit{LLM4CD}, an LLM-augmented framework that overcomes the limitations of traditional ID-based models by extracting semantically rich cognitive connections. Our bi-level encoder framework enhances adaptability to new students and exercises, effectively addressing the cold-start problem in modern intelligent tutoring systems.

\item LLM4CD performs significantly better than existing CD models on four real-world datasets. We also study how different text types provide maximum cognitive information and analyze how LLMs process and interpret these texts. This approach demonstrates how cognitive analysis of texts can be transformed into adequate diagnostic data.
\end{itemize}

\section{RELATED WORK}
\subsection{Cognitive Diagnosis}
Cognitive Diagnosis (CD) is a pivotal component of intelligent tutoring systems (ITS). The CD task is based on the educational assumption that each student's cognitive state is stable in static scenarios. This allows diagnosing students' abilities on specific concepts or skills through their historical interactions~\cite{zhang1997parameter}. Early CD models predominantly adopted two types of modeling approaches: student-item interaction modeling and student-concept interaction modeling. Representing student-item interaction modeling, IRT~\cite{embretson2013item}, MIRT~\cite{ackerman2014multidimensional}, and MF~\cite{koren2009matrix,thai2015multi} utilized manually crafted linear functions to simulate the performance of students on various assessments effectively. For student-concept interaction modeling, DINA~\cite{de2009dina} adopts a discrete analysis approach, evaluating students' mastery over diverse skill attributes and accounting for factors such as slip and guessing, which can influence assessment outcomes. The rise of deep learning~\cite{lecun2015deep} has ushered in advanced modeling techniques, such as NCD~\cite{wang2020neural}, which leverages neural networks to capture higher-order interaction features between students and exercises. More recently, models such as RCD~\cite{gao2021rcd} and SCD~\cite{wang2023self} have incorporated graph neural networks to construct and analyze a comprehensive multi-layered knowledge network encompassing students, exercises, and concepts.

% These CD models, though effective, primarily use identifier (ID) modeling where students, exercises, and concepts are uniquely identified within the system. This method facilitates easy tracking and analysis of interactions. However, relying exclusively on ID-based modeling limits the depth of insights, capturing only explicit interactions and missing nuanced, implicit information in learners' behaviors and educational contexts. 

These CD models, though effective, primarily use ID modeling where students, exercises, and concepts are uniquely identified within the system. While this method facilitates easy tracking and analysis of interactions, it inherently suffers from the cold-start problem when encountering new students or exercises. Recent works have attempted to address this challenge through various approaches, such as transferable knowledge concept graphs~\cite{gao2023leveraging}, zero-shot diagnosis with early-bird students~\cite{gao2024zero}, and inductive learning frameworks~\cite{liu2024inductive}. However, these methods still face limitations in fully representing cognitive states when generalizing to unseen data. Moreover, relying exclusively on ID-based modeling limits the depth of insights, capturing only explicit interactions and missing nuanced, implicit information in learners' behaviors and educational contexts.
% Consequently, there is growing interest in developing more sophisticated models incorporating deeper semantic and relational insights beyond simple IDs to enhance the cognitive diagnostic process.

\subsection{LLM-Enhanced User Modeling}
Recent efforts~\cite{tan2023user,lin2023can,lin2024clickprompt, fu2024sinktstructureawareinductiveknowledge,lin2024rella} have shown that employing LLMs to enhance user modeling has yielded impressive results. Several works~\cite{yin2023heterogeneous, liu2023first} utilize LLMs to develop sophisticated user profiles. The HKFR~\cite{yin2023heterogeneous} system, for example, leverages user heterogeneous behaviors to generate comprehensive profiles via ChatGPT~\cite{ouyang2022training}, encompassing aspects like behavior subjects, content, and scenarios. GENRE~\cite{liu2023first} utilizes LLMs to pinpoint regions and topics of interest based on user browsing history, providing a more personalized content experience. LLMs also play a crucial role from the perspective of feature encoding~\cite{peng2023gpt, runfeng2023lkpnr}. GPT4SM~\cite{peng2023gpt} uses these models to encode recommendation queries and candidate texts to predict relevance more effectively. LKPNR~\cite{runfeng2023lkpnr} employs open-source LLMs like Llama~\cite{touvron2023llama} and RWKV~\cite{peng2023rwkv} for enhanced semantic capture of news content. Moreover, LLMs could offer powerful knowledge augmentation~\cite{xi2023towards, wei2024llmrec}. For instance, KAR~\cite{xi2023towards} utilizes LLMs to generate detailed user and item profiles, enriching the underlying knowledge base for recommendation systems. 
% This capability is extended in systems like PULSAR~\cite{li2023pulsar} and AugESC~\cite{zheng2022augesc}, where LLMs generate contextually enriched dialogues and narrative-driven recommendations, tailoring interactions more closely to user preferences and behaviors. 
% These implementations underscore the transformative impact of LLMs across various facets of user modeling, enabling a deeper, more nuanced understanding of user data and enhancing the overall effectiveness of user interaction systems.

% CD system is a specific application of user modeling within educational systems. 
% Similar to recommendation systems that model user interests, CD models focus on assessing student abilities. 
The CD task differs significantly from other user modeling applications due to its educational context, which is rich with structured information and domain-specific semantic connections. 
Therefore, developing tailored approaches that leverage cognitive perspectives and the deep semantic capabilities of LLMs is essential to integrating textual information in educational settings effectively. 
% This adaptation ensures that CD models accurately reflect complex educational constructs and enhance diagnostic assessments.

\begin{figure*}[t]
  \centering
  \includegraphics[width=\textwidth]{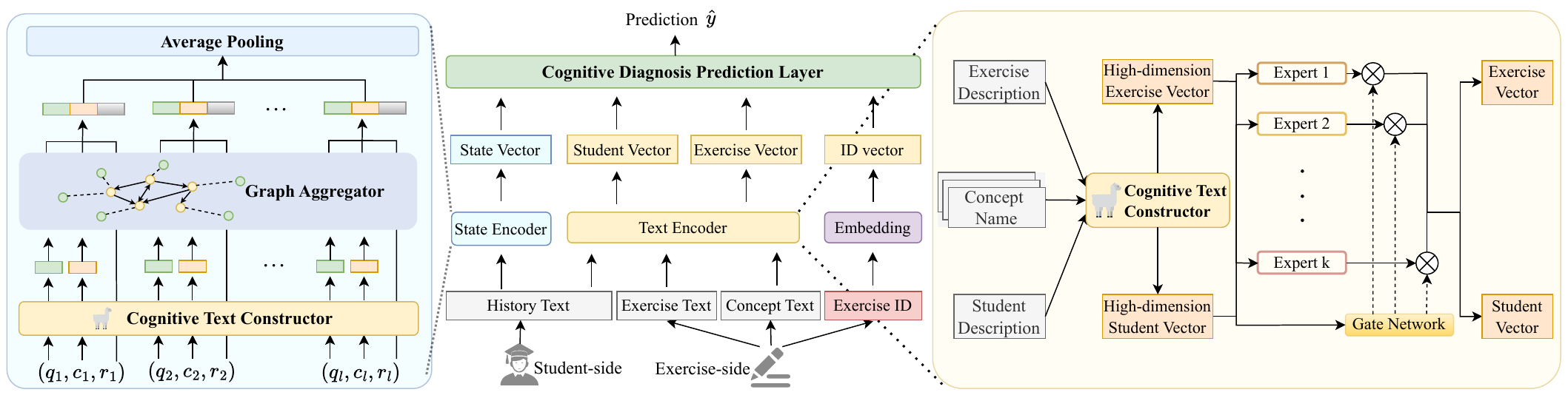} 
  \caption{The overall framework of LLM4CD.}

  \label{fig:lamda}
\end{figure*}
\section{PRELIMINARIES}

\subsection{Problem Formulation}
In ITS, there are $N$ students, $M$ exercises, and $K$ concepts, denoted as $\mathcal{S}=\{s_1,s_2,\cdots, s_N\}$, $\mathcal{Q} = \{q_1, q_2,\cdots,q_M\}$ and $\mathcal{C}=\{c_1, c_2, \cdots, c_K\}$, respectively. 
% $Q$-matrix $Q\in\{0,1\}^{M\times K}$, annotated by experts, indicates the relationships between exercises and concepts, where $Q_{ij}=1$ signifies that exercise $q_i$ is associated with concept $c_j$.

The test history of the student $s\in \mathcal{S}$ is recorded as the set $\mathcal{L}_s = \{(q_i,r_i)\}_{i=1}^{l_s}$, where $q_i\in \mathcal{Q}$ represents the exercise done by the student, and $r_i = \{0,1\}$ indicates the score of student $s$ on exercise $q_i$. 
% \ljh{$\mathcal{L}$ is typically used for loss function, why not $\mathcal{H}$}
In the CD task, given the test history and a new exercise $q$, the model is asked to predict the probability that the student correctly answers the new exercise under his current cognitive state, \textit{i.e.}, $p((q, r=1)\mid \mathcal{L}_s).$

\subsection{Relation Graph}
Benefiting from RCD~\cite{gao2021rcd} and GMOCAT~\cite{wang2023gmocat}, we model the relationships between exercises and concepts within educational contexts using a concept dependency graph and an exercise-concept relationship graph, respectively.

\paragraph{Concept Dependency Graph}

 The concept dependency graph primarily represents the dependencies among concepts. In educational contexts, there are various relationships between concepts, such as prerequisite-successive relationships and correlation relationships. We construct the graph $\mathcal{G}_c(\mathcal{C}, \mathcal{R}_p \cup \mathcal{R}_c)$ to denote prerequisite-successive and correlative relationships between concepts during the learning process of a student, where $\mathcal{C}$ is vertex, $\mathcal{R}_p$ and $\mathcal{R}_c$ denotes the dependency and correlative relationship, respectively.  $\mathcal{R}_p$ and $\mathcal{R}_c$ are generated by LLMs. We will introduce the generation of $\mathcal{R}_p$ and $\mathcal{R}_c$  in Section~\ref{sec:rc}.

\paragraph{Exercise-Concept Relationship Graph}
The exercise-concept relationship graph illustrates the bidirectional relationships between exercises and concepts. We construct the graph $\mathcal{G}_{qc}(\mathcal{Q}\cup\mathcal{C}, \mathcal{R}_{qc})$ based on expert-annotated mappings of exercise-concept pairs, where $\mathcal{Q}, \mathcal{C}$ are vertexes, $\mathcal{R}_{qc}$ denotes the relationships between exercises and concepts.
If exercise $q_i$ is associated with concept $c_j$, there is an edge between $q_i$ and $c_j$ in $\mathcal{G}_{qc}$.
% Specifically, if $r_{q_i\leftrightarrow c_j}=1$, it indicates that exercise $q_i$ primarily involves concept $c_j$.

\paragraph{Relation Graph in LLM4CD}
 We define the comprehensive relation map as $\mathcal{G}(\mathcal{G}c \cup \mathcal{G}{qc}, \mathcal{R}c \cup \mathcal{R}{qc})$, which combines the concept dependency graph and the exercise-concept relationship graph. This integrated graph captures educational dependencies among concepts and the bidirectional correlations between exercises and concepts, allowing for a multi-level representation of educational relationships for our CD model.
% \ljh{?}
\section{METHODOLOGY}
In this section, we will provide a detailed introduction to our method, with the overall framework illustrated in Figure~\ref{fig:lamda}
. Firstly, LLM4CD employs a bi-level encoder architecture to construct cognitively expressive textual representations. The \textit{Text Encoder} generates macro-level cognitive representations from both the student and exercise perspectives, while the \textit{State Encoder} develops micro-level representations of student knowledge states by integrating historical learning data. Secondly, these encoded features are then synergistically combined in the \textit{diagnosis layer} to make the final prediction.

\subsection{Cognitive Text Construction}\label{sec:rc}

\begin{figure*}[!htp]
  \centering
  \includegraphics[width=\textwidth]{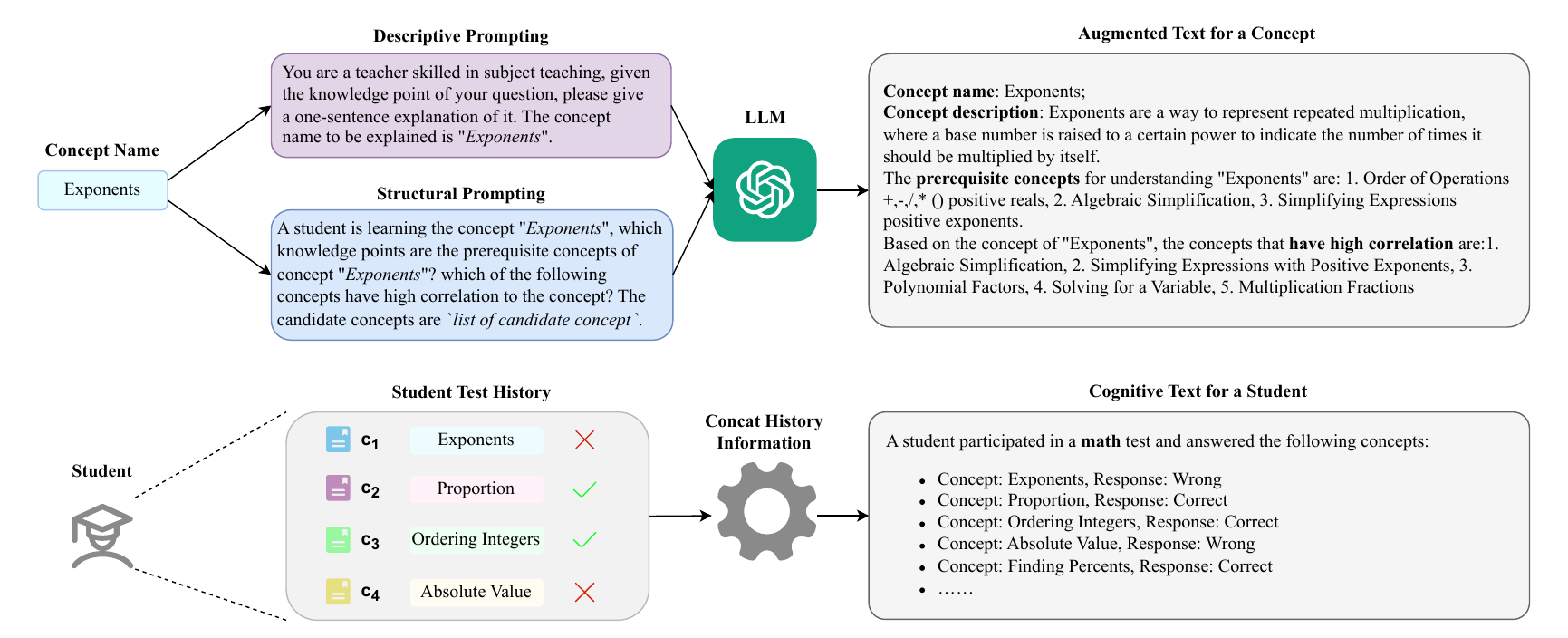} 
  \caption{Text construction from the exercise and student sides. %\dk{Consider: "Rationale Knowledge Prompting" for the top, "Correlated Concepts Prompting" for the bottem. Also, the image is unbalanced, consider to decrease the gap between the purple and blue frames and make the overall heights the same. Maybe highlight the correlated concepts for "exponents" in the final prompt in blue color and highlight the rational knowledge in purple color.}
  }
  \label{fig:text}
\end{figure*}

To better accommodate the logical structure of the quiz scenario and the unique characteristics of the educational dataset, we design cognitive texts from both the exercise and student sides, which serve as inputs for LLM4CD. The process of constructing concept-based and student-based textual representations is illustrated in Figure~\ref{fig:text}.

On the exercise side, we combine the original exercise text \(x^e\) with an augmented description of the relevant concepts to capture the cognitive information embedded in the exercise \(q\). The augmented description of a concept \(c\) includes the concept description \(x_\text{description}\) and the concept structural description \(x_\text{structure}\). Given that concepts in the dataset are often represented by single words, there is potential for ambiguity. For instance, the word "table" could refer to a piece of furniture or a data chart. To address this, we use LLMs to expand and clarify the meaning of the concept texts, thereby generating a more detailed concept description \(x_\text{description}\). Additionally, recognizing the interrelatedness of concepts, we explicitly describe the prerequisite concepts and correlated concepts for the concept \(c\), forming the concept structural description \(x_\text{structure}\). Formally, this process of constructing text for an exercise \(q\) is denoted as:
\begin{equation}
x^q = x^e \oplus x^c,
     x^c = x_\text{description} \oplus x_\text{structure}.
\end{equation}
If the dataset does not contain the exercise text \(x^e\), we directly set \(x^q = x^c.\)

On the student side, due to the lack of explicit descriptions, we construct textual representations by compiling students’ test histories into a comprehensive text sequence \(x^s\). This sequence is generated by concatenating each concept name with the corresponding response, such as "concept: Exponents, response: correct." As illustrated in Figure~\ref{fig:text}, this sequence not only records the students’ answering history but also indirectly reflects their cognitive states concerning various concepts.

After generating cognitive text of concept, exercise, and student, we encode the cognitive texts by an LLM encoder.
We input the cognitive text $x^c$, $x^q$ and $x^s$ into the LLM encoder to acquire their representation vectors $e^c, e^q$, and $e^s$, respectively: 
\begin{equation}
    e^c = \text{LLMEnc}(x^c),  e^q = \text{LLMEnc}(x^q),  e^s = \text{LLMEnc}(x^s).
\end{equation}

\subsection{Text Encoder}
The semantic vectors generated by the LLM possess high-dimensional semantic representations~\cite{li2020sentence}. To adapt these semantic representation vectors for the CD task,  we introduce an adaptor module with a Mixture-of-Experts (MoE) architecture~\cite{jacobs1991adaptive}, which maps the high-dimensional semantic representation vectors $e_q$ and $e_s$ into lower-dimensional vectors, denoted as $h_q$ and $h_s$. Formally, the MoE adaptor can be represented as:
\begin{align}
&\alpha_i^{q} = \text{Softmax}(f(e_{q})),\;& \alpha_i^{s} = \text{Softmax}(f(e_{s})), \\
&h_q^{\text{text}} = \sum_{i=1}^k \alpha_i \times A_i e^q_i,\;& h_s^{\text{text}} = \sum_{i=1}^k \alpha_i \times A_i e^s_i,
\label{eq:moe1}
\end{align}
where $f(\cdot)$ represents the gating networks to calculate the mixture weights, $\alpha_i^q$ and $\alpha_i^s$ are the weights of expert, $A_i$ are learnable matrices to reduce the dimension, $k$ is the hyper-parameter to control the number of experts.

\subsection{State Encoder}
To capture the structural relationships between nodes in the relation graph, we utilize Graph Attention Network~\cite{velickovic2017graph} (GAT) as the graph aggregator to generate embeddings of each node.

\subsubsection{Concept Relationship Aggregation}
For each concept $c$, we encode $\mathcal{R}_p$ and $\mathcal{R}_c$ through aggregating representations of neighbor nodes:
\begin{align}
    g^c_\text{pre} &= \sum\limits_{c_\text{pre}\in \mathcal{N}_c^\text{pre}}\alpha_{c,c_\text{pre}}W_{\text{pre}} e_{c_\text{pre}},\\
    g^c_\text{cor} &= \sum\limits_{c_\text{cor}\in \mathcal{N}_c^\text{cor}}\alpha_{c,c_\text{cor}}W_\text{cor}e_{c_\text{cor}},
\end{align}
where $\mathcal{N}_c^\text{pre}$ and $\mathcal{N}_c^\text{cor}$ represent the sets of neighboring concepts with prerequisite and correlative relationships, respectively, $W_\text{pre}$ and $W_\text{cor}$ are learnable parameter matrices, and $\alpha_{c,c_\text{pre}}$ and $\alpha_{c,c_\text{cor}}$ are attention weights determined by the similarity of embeddings between concept $c$ and its neighbors, calculated as:
\begin{align}
\alpha_{c,c_\text{pre}}&=\text{Softmax}_{c_\text{pre}}\left(\text{ReLU}\left( a_\text{pre}(W_\text{pre}e_c \oplus W_\text{pre}e_{c_\text{pre}})\right)\right),
\label{gat3}\\
\alpha_{c,c_\text{cor}}&=\text{Softmax}_{c_\text{cor}}\left(\text{ReLU}\left(a_\text{cor}(W_\text{cor}e_c \oplus W_\text{cor}e_{c_\text{cor}})\right)\right),
\label{gat4}
\end{align}
where $a_\text{pre}$ and $a_\text{cor}$ represent a linear layer, $\oplus$ indicates concatenate operation.

We adopt an attention mechanism to model the preference between prerequisite and correlative relations:
\begin{align}
w^c_\text{pre} &= \text{ReLU}\left(W'_\text{pre} g_\text{pre}^c + b'_\text{pre}\right),\\
    w^c_\text{cor} &= \text{ReLU}\left(W'_\text{cor}  g_\text{cor} + b'_\text{cor}\right),
\label{gat5}\\
[\mu_\text{pre}, \mu_\text{cor}] &= \text{Softmax}\left([w_\text{pre}, w_\text{cor}]\right),
\label{gat7}
\end{align}
where $W'_\text{pre}$, $W'_\text{cor}$, $b'_\text{pre}$ and $b'_\text{cor}$ are learnable parameters.
The aggregated embedding for the concept $c$ could be written as
\begin{equation}
\tilde{e}^c = e_c+\mu_\text{pre}g_\text{pre}+\mu_\text{cor}g_\text{cor}.
\label{gat8}
\end{equation}

\subsubsection{Exercise-Concept Relationship Aggregation}
Similar to the concept relationship aggregation, we aggregate the embeddings of neighbors in the exercise-concept relationship graph with attention weights:
\begin{gather}
g_\text{rel} = \sum\limits_{c'\in N_q^\text{rel}}\beta_{q,c'}W_\text{rel}e_{c'},
\label{gat7}\\
    \beta_{q,c'} = \text{Softmax}_{c'} \left( \text{ReLU} \left( a_\text{rel} \left(W_\text{rel}e_c \oplus W_\text{rel} e_{c'}\right)\right)\right),
\end{gather}
where $\mathcal{N}_q^\text{rel}$ as the set of neighboring concept nodes for exercise $q$ in the exercise-concept relationship graph.
The representation of the exercise is
\begin{equation}
\tilde{e}^q = g_\text{rel}+e_q.
\label{gat8}
\end{equation}
 % to produce a correlation-aware embedding $h_\text{rel}$:

\subsubsection{State Aggregation}
To capture a student's knowledge state comprehensively, we integrate relation-aware embeddings $\tilde{e}_c$ and $\tilde{e}_q$ with the  response of the student. 
For a record $(q, r)$, we encode the response $r$ through an embedding layer $\text{E}_r$ and get its representation vector $e^r = \text{E}_r(r).$
And each record $(q_i,r_i)$ could be represented by 
\begin{equation}
    h_i = \tilde{e}^{q}_i\oplus \tilde e^{c}_i\oplus e^{r}_i
\end{equation}
where $c_i$ is the concept related to the exercise $q_i$.

When student taking an exam, his knowledge state  remains unchanged, \textit{i.e.,} the order of previous exercises does not influence his answer to the next question. Therefore, we choose average pooling to aggregate the test history of the student $s$:
\begin{equation}
h_s^\text{state} = \sigma\left(W\left(\frac{1}{l_s}\sum\limits_{i=1}^{l_s}h_i\right)+b\right),
\label{att3}
\end{equation}
where $W,b$ are trainable parameters, $\sigma$ is Sigmoid function.
% This streamlined process ensures a refined, aggregated representation of the student's knowledge state.

\subsection{Diagnosis Layer}
Inspired by NCD~\cite{wang2020neural}, we use a feed forward layer to integrate student state and current exercise and make the  prediction of the CD task.
On the student side, we fuse semantic and  state representations to represent estimated knowledge state of the student $s$:
\begin{equation}
h_s = h_s^\text{text} +h_s^\text{state}.
\label{cd1}
\end{equation}
On the exercise side, since some datasets does not have exercise texts and multiple exercises may share the same concept, we incorporate learnable ID embeddings $\text{E}_q$ for exercises to ensure the model can make predictions to specific exercise:
\begin{equation}
    h_q = h_q^\text{text} + \text{E}_q(q).
\end{equation}
When exercise texts are available, LLM4CD supports training on plain text. Specific experimental results are demonstrated in RQ3.

Following the established practice in previous CD models~\cite{wang2020neural,gao2021rcd}, we model the interaction between student proficiency and exercise difficulty through their difference. Specifically, we compute the diagnostic feature vector as the difference between student proficiency representation $h_s$ and exercise difficulty representation $h_q$:
\begin{equation}
\hat{y} = \sigma\left(F_\text{predict}\left(h_s - h_q\right)\right).
\label{cd3}
\end{equation}
Here, $\hat{y}$ predicts the probability that student $s$ correctly answers exercise $q$, $F_\text{predict}$ is a fully connected prediction layer with a ReLU activation and dropout strategy. Ultimately, we use a cross-entropy loss function as loss function:
\begin{equation}
\mathcal{L}=-\sum\limits_{i}\left(r_i\log{\hat{y}_i}+\left(1-r_i\right)\log\left(1-\hat{y}_i\right)\right).
\label{cd4}
\end{equation}

\section{EXPERIMENTS}
To validate the effectiveness of LLM4CD, we conduct experiments on four educational datasets. These experiments are designed to answer the following five research questions (RQs):

\begin{itemize}
    \item[\textbf{RQ1}] How does the proposed LLM4CD perform compared to the state-of-the-art CD models?
    \item[\textbf{RQ2}] Does the introduction of semantic information enhance the diagnostic capabilities of existing CD models?
    \item[\textbf{RQ3}] How does LLM4CD cope with the cold-start problem where new students and exercises come into the CD task?
    \item[\textbf{RQ4}] How do different text representations affect the performance of LLM4CD?
    \item[\textbf{RQ5}] How do different components of LLM4CD influence its performance?
    \item[\textbf{RQ6}] How about the interpretation of LLM4CD on diagnosing student knowledge states for CD?
\end{itemize}

\subsection{Experimental settings}
\subsubsection{Datasets}
In our experiments, we evaluate our method on four datasets:
(1) ASSIST09\footnote{https://sites.google.com/site/assistmentsdata/home/2009-2010-assistment-data?authuser=0}, (2)ASSIST12\footnote{https://sites.google.com/site/ASSISTdata/home/2012-13-school-data-with-affect}, (3) Junyi Academy Dataset\footnote{https://pslcdatashop.web.cmu.edu/DatasetInfo?datasetId=1198}, and (4) Programming:

% ASSIST09, ASSIST12, and Junyi are public datasets focused on mathematics, widely used in intelligent education research. The Programming dataset, which includes exercise texts, is a private dataset used in programming education. \dk{You could delete this paragraph if space is limited, since the detailed descriptions are offered below.}

% The basic statistics of these four datasets are listed in Table~\ref{tab:dataset} as descriptions are as follows: \dk{Remove this paragraph.}

\begin{itemize}[leftmargin=10pt]
    \item The ASSIST09 dataset was ollected from an online mathematics tutoring platform during the 2009-2010 school year, this dataset includes concept texts but lacks exercise descriptions.
    \item Similar to ASSIST09 dataset, ASSIST12 was also collected during the 2012-2013 school year with concept texts but no exercise descriptions.
    \item The Junyi Academy Dataset was collected from a Taiwanese e-learning platform (2010-2015), uses KC-Exercise (Knowledge Components) as concept text, but lacks exercise texts.
    \item The Programming dataset was collected from a commercial platform (2021-2023), includes both concept texts and exercise descriptions.
\end{itemize}

ASSIST09, ASSIST12, and Junyi are among the most well-known and widely-used public benchmarks in intelligent education research, focusing on mathematics education and serving as standard evaluation datasets for CD models. Additionally, we include the Programming dataset, which is a private dataset used in programming education that uniquely contains exercise texts.

The detailed statistics of the four datasets are listed in Table~\ref{tab:dataset}.

% \usepackage{graphicx}
\begin{table}[]
\renewcommand{\arraystretch}{1.2} % 
\centering
\caption{Dataset Statistics.}
\label{tab:dataset}
\resizebox{\columnwidth}{!}{%
\begin{tabular}{ccccc}
\hline\toprule
\textbf{Datasets} & \textbf{ASSIST09} & \textbf{ASSIST12} & \textbf{Junyi} & \textbf{Programming} \\ \hline
Subjects        & Math    & Math    & Math    & Programming \\
\#Students      & 1088    & 2550    & 3118    & 2337        \\
\#Exercises     & 15279   & 45706   & 2791    & 1256        \\
\#Concepts      & 136     & 245     & 723     & 116         \\
\#Interactions  & 139909  & 391549  & 495584  & 210118      \\
Attempts per s. & 129     & 154     & 159     & 90          \\
Correct Rate    & 68.18\% & 65.86\% & 54.05\% & 73.24\%     \\ \hline\toprule
\end{tabular}%
}
\end{table}

\begin{table*}[]
\renewcommand{\arraystretch}{1.2} % 
\caption{Overall performance of LLM4CD and baselines on four real-world datasets. LLM4CD-C and LLM4CD-V choose ChatGLM2 and Vicuna as LLM encoders, respectively. All the Text-based methods introduce semantic features encoded by ChatGLM2. Existing state-of-the-art results are underlined, and the best results are bold. * indicates p-value $< 0.05$ in the t-test.  }

\label{tab:main}
\resizebox{\textwidth}{!}{%
\begin{tabular}{cc|ccc|ccc|ccc|ccc}
\hline\toprule
\multicolumn{2}{c|}{\multirow{2}{*}{\textbf{Models}}} &
  \multicolumn{3}{c|}{\textbf{ASSIST09}} &
  \multicolumn{3}{c|}{\textbf{ASSIST12}} &
  \multicolumn{3}{c|}{\textbf{Junyi}} &
  \multicolumn{3}{c}{\textbf{Programming}} \\ \cline{3-14} 
\multicolumn{2}{c|}{} &
  \textbf{AUC} &
  \textbf{ACC} &
  \textbf{RMSE} &
  \textbf{AUC} &
  \textbf{ACC} &
  \textbf{RMSE} &
  \textbf{AUC} &
  \textbf{ACC} &
  \textbf{RMSE} &
  \textbf{AUC} &
  \textbf{ACC} &
  \textbf{RMSE} \\ \hline
\multicolumn{1}{c|}{\multirow{6}{*}{Baseline}}   & DINA   & 0.6945 & 0.7002 & 0.4985 & 0.6743 & 0.6949 & 0.4506 & 0.7088 & 0.6855 & 0.4981 & 0.7364 & 0.7507 & 0.4107 \\
\multicolumn{1}{c|}{}                            & IRT    & 0.7116 & 0.7019 & 0.4484 & 0.7130 & 0.7200 & 0.4387 & 0.7816 & 0.7167 & 0.4434 & 0.7810 & 0.7680 & 0.3956 \\
\multicolumn{1}{c|}{}                            & MIRT   & 0.7572 & 0.7225 & 0.4369 & 0.7199 & 0.7189 & 0.4395 & 0.8118 & 0.7378 & 0.4194 & 0.7650 & 0.7587 & 0.4031 \\
\multicolumn{1}{c|}{}                            & MF     & 0.7330 & 0.7111 & 0.4389 & 0.7224 & 0.7188 & 0.4386 & 0.8094 & 0.7358 & 0.4206 & 0.7777 & 0.7658 & 0.3976 \\
\multicolumn{1}{c|}{}                            & NCD    & 0.7400 & 0.7175 & 0.4415 & 0.7291 & 0.7181 & 0.4397 & 0.8061 & 0.7328 & 0.4248 & 0.7428 & 0.7474 & 0.4115 \\
\multicolumn{1}{c|}{}                            & RCD    & 0.7652 & 0.7313 & 0.4235 & 0.7434 & 0.7321 & 0.4259 & 0.8137 & 0.7388 & 0.4182 & 0.7793 & 0.7665 & 0.4182 \\ \hline
\multicolumn{1}{c|}{\multirow{4}{*}{Text Augmented}} & T-MIRT & 0.7573 & 0.7228 & 0.4316 & 0.7275 & 0.7239 & 0.4352 & 0.8156 & 0.7404 & 0.4175 & 0.7728 & 0.7629 & 0.4012 \\
\multicolumn{1}{c|}{}                            & T-MF   & 0.7521 & 0.7229 & 0.4319 & 0.7298 & 0.7251 & 0.4344 & 0.8125 & 0.7370 & 0.4192 & 0.7773 & 0.7660 & 0.3979 \\
\multicolumn{1}{c|}{}                            & T-NCD  & 0.7486 & 0.7261 & 0.4332 & 0.7349 & 0.7289 & 0.4354 & 0.8116 & 0.7371 & 0.4201 & 0.7742 & 0.7622 & 0.3996 \\
\multicolumn{1}{c|}{}                            & T-RCD  & 0.7652 & 0.7328 & 0.4236 & 0.7387 & 0.7293 & 0.4293 & 0.8149 & 0.7389 & 0.4175 & 0.7831 & 0.7674 & 0.3951 \\ \hline
\multicolumn{1}{c|}{\multirow{2}{*}{Ours}} &
  LLM4CD-C &
  \textbf{0.7788}* &
  \textbf{0.7397}* &
  \textbf{0.4182}* &
  \textbf{0.7508}* &
  \textbf{0.7352}* &
  \textbf{0.4241}* &
  \underline{0.8167} &
  \underline{0.7412} &
  \textbf{0.4171}* &
  \underline{0.7882} &
  \textbf{0.7712}* &
  \underline{0.3937} \\
\multicolumn{1}{c|}{} &
  LLM4CD-V &
  \underline{0.7769} &
  \underline{0.7381} &
  \underline{0.4188} &
  \underline{0.7468} &
  \underline{0.7335} &
  \underline{0.4248} &
  \textbf{0.8185}* &
  \textbf{0.7422}* &
  \underline{0.4173} &
  \textbf{0.7900}* &
  \underline{0.7740} &
  \textbf{0.3932}* \\ \hline\toprule
\end{tabular}%
}
\end{table*}

\subsubsection{Compared Methods}
We evaluate LLM4CD against several representative CD methods.
% {IRT}~\cite{embretson2013item}, {MIRT}~\cite{ackerman2014multidimensional},{MF}~\cite{koren2009matrix} and {DINA}~\cite{de2009dina} are traditional CD models that use parameters to model student-item or student-concept interactions.
% {NCD}~\cite{wang2020neural} and {RCD}~\cite{gao2021rcd} are  the latest CD models that utilize deep learning.
\begin{itemize}[leftmargin=10pt]
\item \textbf{IRT}~\cite{embretson2013item} uses linear functions to model students' unidimensional abilities and the items' characteristics.
\item  \textbf{MIRT}~\cite{ackerman2014multidimensional} models multiple dimensions of student knowledge and item characteristics.
\item  \textbf{DINA}~\cite{de2009dina} assumes that student skill mastery is a deterministic "and" gate process, incorporating noise to account for random errors.
\item  \textbf{MF}~\cite{koren2009matrix} decomposes the matrix of student response scores to extract latent feature vectors for students and items, predicting student performance.
\item  \textbf{NCD}~\cite{wang2020neural} is a deep learning-based CD model that employs neural networks to model complex higher-order student interactions.
\item \textbf{RCD}~\cite{gao2021rcd} uses relation maps to model relationships among concepts and between students and items, integrating these maps into the CD task.
\end{itemize}

To explore the performance of different LLMs in the CD task, we also deploy Sentence-BERT~\cite{reimers2019sentence}, ChatGLM2-6B ~\cite{du2021glm}, and Vicuna-7B~\cite{chiang2023vicuna} as LLM encoders for comparison.
These models are widely used for extracting embeddings in various user modeling tasks.

\subsubsection{Parameter Settings and Metrics}

All trainable parameters in our model are initialized using the Xavier method~\cite{nakagawa2019graph} to ensure stable gradient propagation. We optimize model parameters with the AdamW optimizer~\cite{loshchilov2018fixing}. The learning rate is chosen from $\{0.001, 0.005, 0.0001\}$ with a weight decay at each epoch. The batch size during training is selected from $\{400, 1000, 4000\}$. 
% \dk{Did we search the optimal hyperparameters? If so, list the search range. For example, the learning rate is searched in the range \{0.001, 0.005, 0.01\}. Otherwise, the current statements show that we did not do parameter searching.}

Embedding dimensionality is set to match the number of concepts in the dataset. The dataset is split into training (70\%), validation (10\%), and test (20\%) subsets. We evaluate the model using AUC, accuracy (ACC), and root mean square error (RMSE) metrics, retaining the model parameters that achieve the best validation performance. The source code is open-sourced in our repository. \footnote{Source code: \url{https://github.com/yevzh/LLM4CD-Release}}. 

\subsubsection{Deployment Efficiency}

In practical deployment scenarios, inference efficiency is crucial for ensuring real-time responsiveness in ITS. In LLM4CD, we address this by precomputing embeddings for exercises and concepts during a preprocessing step. At inference time, only the student's interaction data needs to be processed alongside these precomputed embeddings. The average inference time per query is approximately 0.28 s/query. This pipeline ensures that our LLM4CD meets the real-time requirements of modern ITS applications.

\subsection{Overall Performance (RQ1)}

\begin{figure*}[t]
    \centering
    \begin{subfigure}{.23\linewidth}
        \centering
        \includegraphics[width=\linewidth]{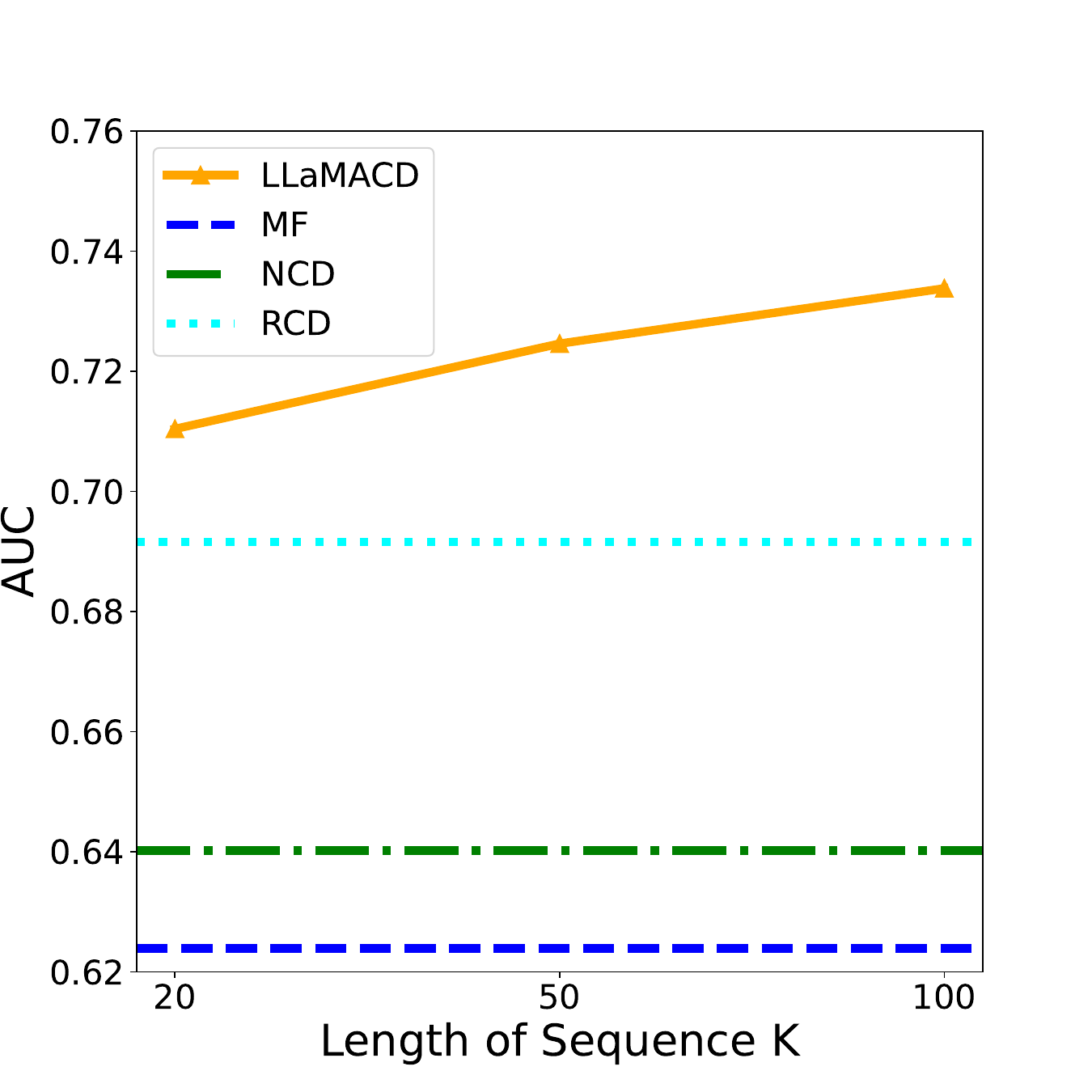}
        \caption{ASSIST09.}
    \end{subfigure}
    \begin{subfigure}{.23\linewidth}
        \centering
        \includegraphics[width=\linewidth]{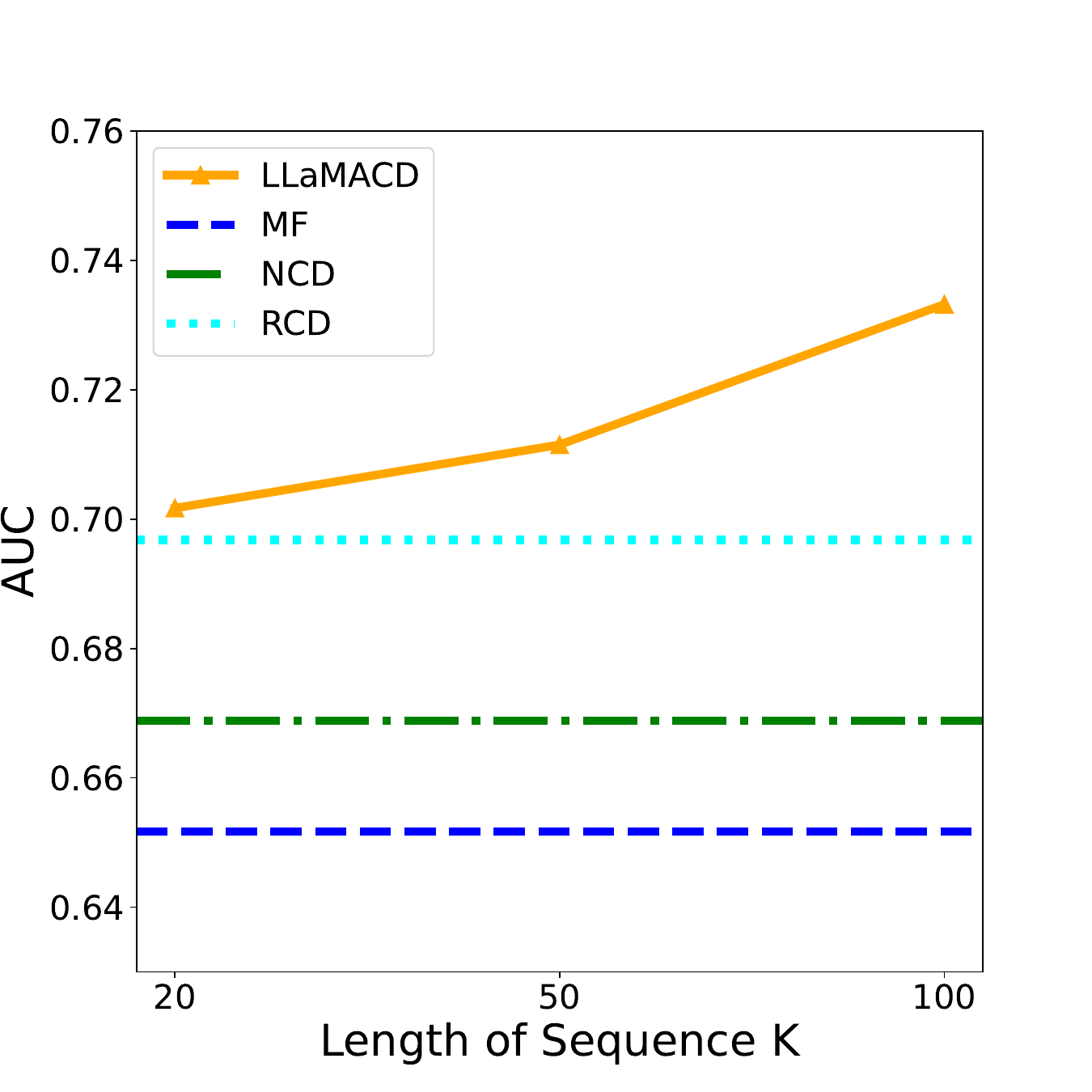}
        \caption{ASSIST12.}
    \end{subfigure}
    \begin{subfigure}{.23\linewidth}
        \centering
        \includegraphics[width=\linewidth]{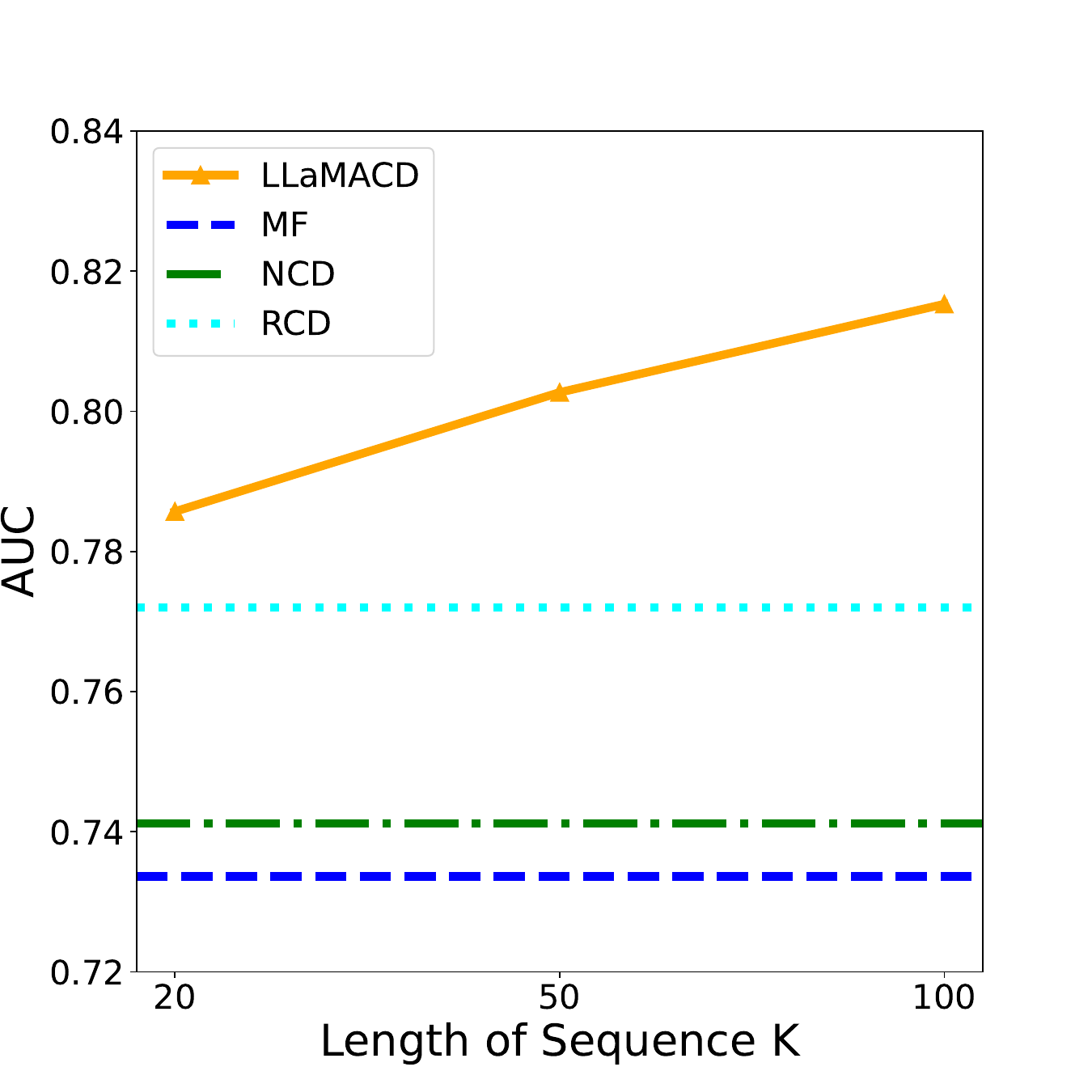}
        \caption{Junyi.}
    \end{subfigure}
    \begin{subfigure}{.23\linewidth}
        \centering
        \includegraphics[width=\linewidth]{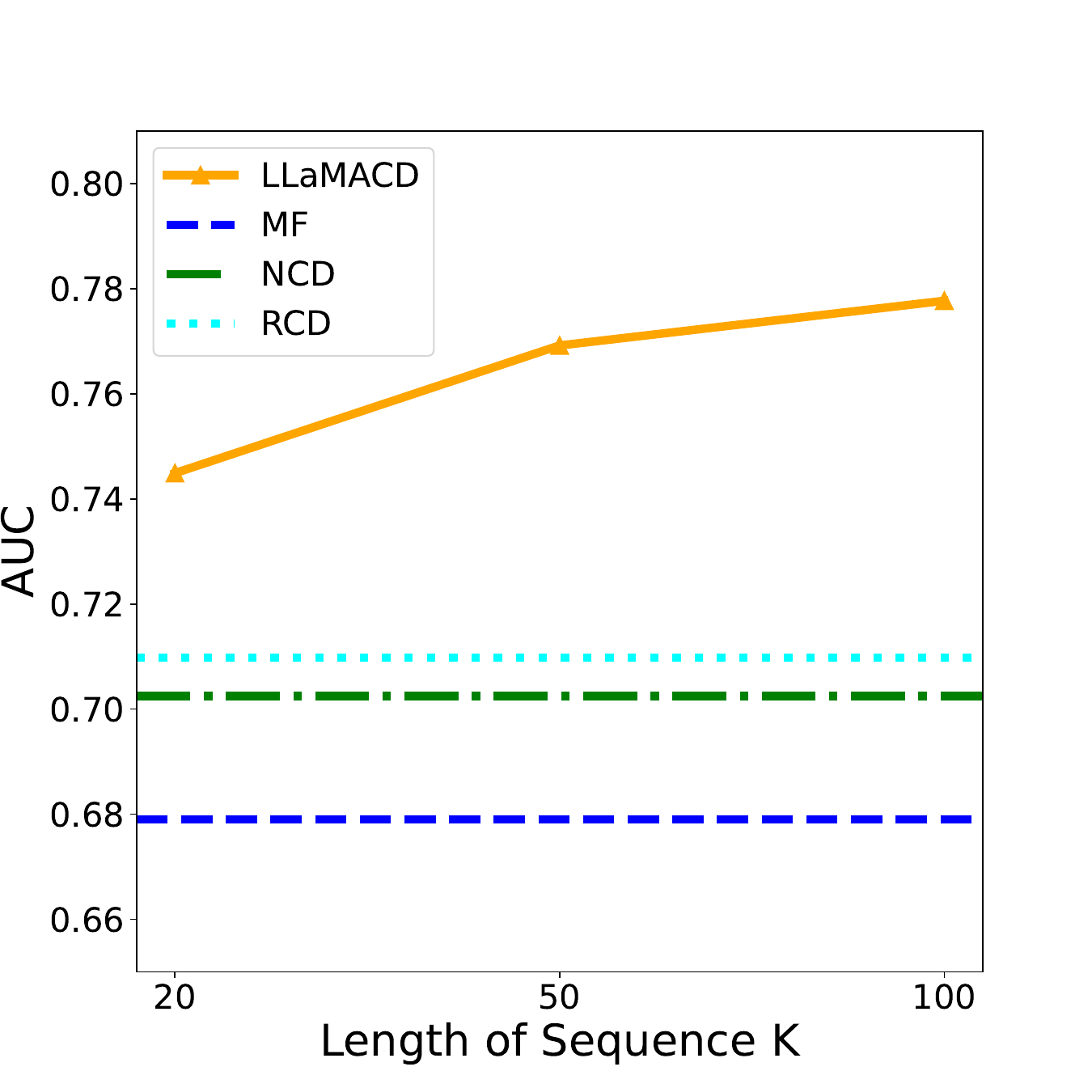}
        \caption{Programming.}
    \end{subfigure}
    \vspace{-5pt}
    \caption{Performance on cold-start prediction for new students in CD models on four datasets. The Length of Sequence $K$ represents the number of interactions per student that have not been seen during training.}
    \vspace{-8pt}
    \label{fig:newstudent}
\end{figure*}

We conduct comparative experiments on four datasets to verify the performance of our proposed LLM4CD framework. Two different LLMs are used as semantic encoders: ChatGLM2 (LLM4CD-C) and Vicuna (LLM4CD-V). The overall performance is shown in Table~\ref{tab:main}, from which we can draw the following observations:

(1) Our LLM4CD model outperforms all baselines, with an average improvement of 1.07\% on AUC over the previously best-performing model (RCD). This demonstrates that LLM4CD effectively integrates open-world knowledge and semantic information to enhance CD tasks.

(2) ChatGLM2 demonstrates superior encoding performance on the ASSIST09 and ASSIST12 datasets compared to Vicuna, whereas Vicuna outperforms ChatGLM2 on the other two datasets. This variance suggests that different LLMs may exhibit diverse semantic understanding capabilities due to their distinct training corpora and methodologies.

(3) On datasets with sparse ID embeddings, such as ASSIST09 and ASSIST12, the integration of textual information provides more significant performance improvements. In contrast, the Junyi dataset shows limited improvement, as its ID embeddings are already dense and provide sufficient representation, leaving less room for textual information to further enhance the model's performance.

\begin{table}[t]
\caption{Performance on cold-start prediction for new students and new exercises on Programming dataset with exercise text.}
\label{tab:cold}
\renewcommand{\arraystretch}{1.1}  % Increase the row spacing
\resizebox{\columnwidth}{!}{%
\begin{tabular}{c|l|cccc}

\hline\toprule
\multirow{2}{*}{\textbf{Cold-start Type}} & \multicolumn{1}{c|}{\multirow{2}{*}{\textbf{Metrics}}} & \multicolumn{4}{c}{\textbf{Models}} \\ \cline{3-6} 
                               & \multicolumn{1}{c|}{} & \textbf{LLM4CD} & \textbf{RCD} & \textbf{NCD} & \textbf{MF} \\ \hline
\multirow{2}{*}{New Students}  & AUC                   & 0.7692           & 0.7098       & 0.7025       & 0.679       \\ \cline{2-6} 
                               & ACC                   & 0.7458           & 0.7272       & 0.7204       & 0.6937      \\ \hline
\multirow{2}{*}{New Exercises} & AUC                   & 0.6746           & 0.6388       & 0.5719       & 0.598       \\ \cline{2-6} 
                               & ACC                   & 0.7341           & 0.7109       & 0.7208       & 0.6035      \\ \hline\toprule
\end{tabular}%
}
\end{table}

\subsection{The Effectiveness of Semantic Embeddings (RQ2)}

To investigate whether semantic vectors can enhance CD capabilities, we integrate the semantic vectors generated by the text encoder into current CD models (MIRT, MF, NCD, RCD). Specifically, while current CD models rely on ID embeddings, we combine these ID embeddings with semantic vectors to create new, enriched representations within these models. 
As shown in Table~\ref{tab:main}, adding semantic vectors could improve the performance of MF and NCD, but does not significantly benefit the RCD model. This is because the RCD model's structure could not directly integrate semantic information. Additionally, smaller models like Sentence-BERT, although effective in general-purpose tasks, demonstrate limited capacity to capture the intricate semantic and logical relationships required in CD tasks. Therefore, we focus on larger LLMs, which are better suited to fully leverage the proposed architecture and meet the demands of this specific domain.

In summary, effective use of semantic vectors requires a well-designed model structure to aggregate the information properly. In the LLM4CD framework, the text encoder aligns the semantic space with the CD task, enabling efficient utilization of semantic information.

\subsection{Cold-Start Analysis of New Students and Exercises (RQ3)}

% We assess LLM4CD's performance on new students unseen in the training dataset. These students have some interaction records but are not included in the training process. We use 20, 50, and 100 interaction records per student as test histories for prediction, eliminating the need for student ID embeddings. We compare LLM4CD's performance with three ID-based models, which use random embeddings for students in the test set. As shown in Figure~\ref{fig:newstudent}, LLM4CD demonstrates its superior capability of modeling student cognition through text representation, leading to more accurate assessments. The results also indicate that longer student test histories further enhance the model's ability to represent cognitive states accurately.

To evaluate LLM4CD's effectiveness in addressing the cold-start problem, we first assess its performance on students who are completely unseen in the training dataset. These students have some interaction records but are not included in the training process, representing a typical cold-start scenario in real-world applications. We use 20, 50, and 100 interaction records per student as test histories for prediction, eliminating the need for student ID embeddings. We compare LLM4CD's cold-start performance with three ID-based models, which use random embeddings for students in the test set. As shown in Figure~\ref{fig:newstudent}, LLM4CD demonstrates its superior capability in handling cold-start scenarios through text representation, leading to more accurate assessments. The results also indicate that longer student test histories further enhance the model's ability to represent cognitive states accurately under cold-start conditions.

% \usepackage{multirow}
% \usepackage{graphicx}

Given that the other three datasets lack exercise text, we evaluate the generalization ability to new exercises of LLM4CD on the Programming dataset, which includes exercise texts. 
To address the issue of lacking historical records for the new exercise, we eliminate the ID input and use text information as the sole feature of the items.
% Given the availability of exercise texts, LLM4CD can rely solely on text-based modeling, operating in an ID-free manner. 
We set aside 20\% of the exercises as a test set, with the remaining 80\% used for training, excluding interactions with the test exercises during training. We then compare its performance with ID-based models. Table~\ref{tab:cold} shows that LLM4CD significantly outperforms ID-only models, demonstrating superior adaptability and performance of LLM4CD in cold-start predictions for new exercises.

\subsection{Text Representation Analysis (RQ4)}

\begin{figure}[t]
  \centering
  \includegraphics[width=0.9\linewidth]{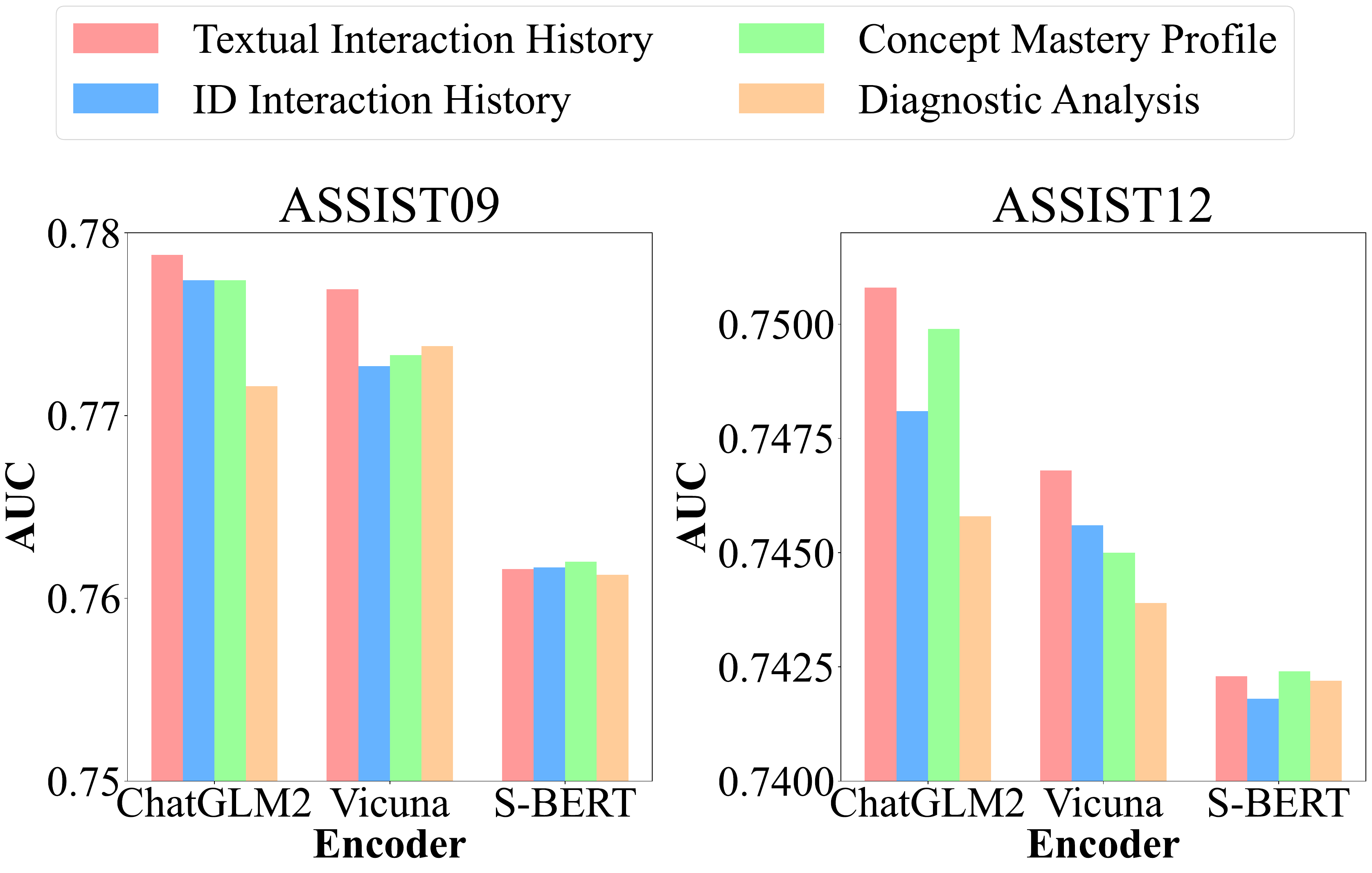} 
  % \caption{Student side's cognitive text representation analysis of LLM4CD.}
  \caption{Comparison of different text representations from the student side.}
  \label{fig:rq2_stu}
\end{figure}

\begin{figure}[t]
  \centering
  \includegraphics[width=0.9\linewidth]{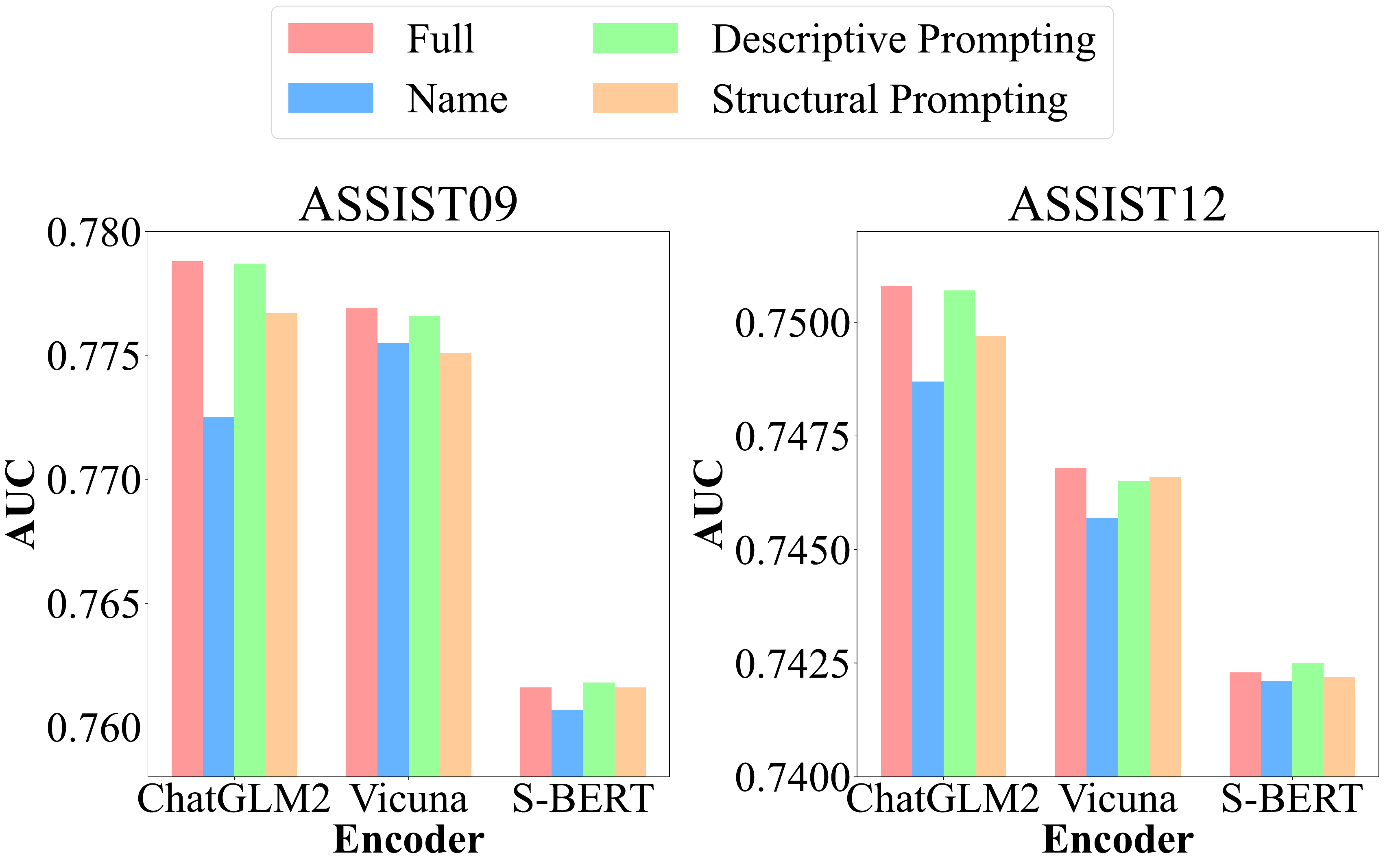} 
  \caption{Comparison of different text representations from the exercise side.}
  \label{fig:rq2_exer}
\end{figure}

We conduct a series of experiments from a cognitive perspective to analyze the impact of different text types and the encoding capabilities of various LLMs in the context of CD. We utilize widely recognized LLMs, including ChatGLM2, Vicuna, and the smaller model, Sentence-BERT. 

The LLMs under study, each with its unique pre-training corpora and model sizes, exhibit diverse comprehension capabilities across different text types. In the context of student modeling, we have developed four unique forms of cognitive texts: 

\begin{itemize}[leftmargin=10pt]
\item \textbf{Textual Interaction History} concatenates students' responses in the "concept text-response" format, creating a continuous text that represents the student's answering process.
\item \textbf{ID Interaction History} forms a long text by concatenating "concept ID-response," summarizing the student's answering activities.
\item \textbf{Concept Mastery Profile} constructs a profile by pairing concept descriptions with corresponding response statistics to reflect the student's mastery of various concepts.
\item \textbf{Diagnostic Analysis} uses GPT-3.5-Turbo~\cite{ouyang2022training} to analyze the student's concept interaction profile and generate a comprehensive diagnostic analysis.
\end{itemize}
This approach enables us to evaluate the impact of text construction on semantic integration in CD models. Additionally, it provides valuable insights into the understanding of LLMs within educational environments.
% \dk{Textual Interaction History, ID Interaction History, Concept Mastery Profile, Diagnostic Analysis}

We conduct experiments on the ASSIST09 and ASSIST12 datasets. As illustrated in Figure~\ref{fig:rq2_stu}, encoding with textual interaction history yields the best results on both datasets. This is due to the rich and lengthy nature of the text, which allows LLMs to more accurately capture students' cognitive states. Encoding using concept names is more effective than ID-based profiles, as it leverages latent semantic relationships to better reflect knowledge in an open-world context.

Sentence-BERT demonstrates minimal variation in encoding performance across different textual representations. This limited performance is likely due to Sentence-BERT's restricted parameter capacity, which affects its ability to process long texts effectively, leading to less precise semantic information. Consequently, its feature representations are considerably less effective compared to those produced by LLMs. This reinforces the superior capability of LLMs in CD tasks, highlighting their advanced ability to understand and encode complex knowledge.

In addition to examining student-side text construction, we also explore the effects of enhancements on exercise-side text. We compare our approach with methods that use only concept names, descriptive enhancements, and structural enhancements. The results, presented in Figure~\ref{fig:rq2_exer}, indicate that incorporating open-world knowledge from both perspectives significantly enhances the quality of semantic features in the CD model, validating the effectiveness of our enhancements.
% \dk{Concretely, we compare our approach with methods that use only the concept name, descriptive enhancements, and structural enhancements, respectively. The results, shown in Figure~\ref{fig:rq2_exer}, indicate that  xxx}

\subsection{Ablation Study (RQ5)}

\begin{table}[t]
\caption{Ablation study of LLM4CD on all datasets.}
\label{tab:ablation}
\renewcommand{\arraystretch}{1.5}  % Increase the row spacing
\resizebox{\columnwidth}{!}{%
\begin{tabular}{l|ll|ll|ll|ll}
\hline\toprule
\multicolumn{1}{c|}{\multirow{2}{*}{\textbf{Models}}} &
  \multicolumn{2}{c|}{\textbf{ASSIST09}} &
  \multicolumn{2}{c|}{\textbf{ASSIST12}} &
  \multicolumn{2}{c|}{\textbf{Junyi}} &
  \multicolumn{2}{c}{\textbf{Programming}} \\ \cline{2-9} 
\multicolumn{1}{c|}{} &
  \multicolumn{1}{c}{\textbf{AUC}} &
  \multicolumn{1}{c|}{\textbf{ACC}} &
  \multicolumn{1}{c}{\textbf{AUC}} &
  \multicolumn{1}{c|}{\textbf{ACC}} &
  \multicolumn{1}{c}{\textbf{AUC}} &
  \multicolumn{1}{c|}{\textbf{ACC}} &
  \multicolumn{1}{c}{\textbf{AUC}} &
  \multicolumn{1}{c}{\textbf{ACC}} \\ \hline
LLM4CD       & 0.7788 & 0.7397 & 0.7508 & 0.7352 & 0.8185 & 0.7422 & 0.7900 & 0.7740 \\
LLM4CD-State & 0.7750 & 0.7397 & 0.7503 & 0.7300 & 0.8175 & 0.7406 & 0.7892 & 0.7727 \\
LLM4CD-Text  & 0.7668 & 0.7287 & 0.7423 & 0.7308 & 0.8164 & 0.7388 & 0.7855 & 0.7701 \\
LLM4CD-LLM   & 0.7504 & 0.7210 & 0.7343 & 0.7288 & 0.8149 & 0.7385 & 0.7782 & 0.7632 \\
LLM4CD-GAT   & 0.7720 & 0.7341 & 0.7498 & 0.7366 & 0.8177 & 0.7412 & 0.7892 & 0.7733 \\
LLM4CD-MoE   & 0.7614 & 0.7317 & 0.7325 & 0.7250 & 0.8165 & 0.7400 & 0.7882 & 0.7693 \\ \hline\toprule
\end{tabular}%
}
\end{table}

We conduct an ablation study involving four experimental variants to investigate the impact of different components in our LLM4CD model. Each variant omits specific parts of the original LLM4CD framework to isolate and identify the contributions of individual components to the overall model performance:
\begin{itemize}[leftmargin=10pt]
    \item \textbf{LLM4CD-State} omits the fine-grained historical record graph embeddings of student knowledge states.
    \item \textbf{LLM4CD-Text} excludes the template-based macro text embeddings from the student side.
    \item \textbf{LLM4CD-LLM} replaces the text representations with ID embeddings.
    \item \textbf{LLM4CD-GAT} replaces the GAT layer in the state encoder with a Linear layer.
    \item \textbf{LLM4CD-MoE} replaces the MoE adaptor in the text encoder with two linear layers for students and exercises respectively.
\end{itemize}

We can draw the following conclusions from the results presented in Table~\ref{tab:ablation}. Firstly, replacing the textual semantic embeddings derived from LLMs in LLM4CD with ID embeddings results in a substantial decrease in performance. This underscores the critical role of textual semantic embeddings generated by LLMs in enhancing the accuracy of CD tasks. Secondly, removing either the macro-level text encoder or the micro-level state encoder for student modeling adversely affects performance. Among these, the macro-level text encoder plays a more vital role in CD tasks, highlighting its importance in comprehensively assessing student cognitive states. Thirdly, substituting the GAT with a linear layer has a minor impact on performance, indicating that the linear layer's aggregation capability is less effective than the GAT in the state encoder. Furthermore, replacing the MoE adaptor for semantic embedding extraction with linear layers that separately extract information for students and exercises results in diminished effectiveness of semantic information extraction, suggesting that our MoE mechanism is more proficient in this regard.

\subsection{Diagnostic Report Analysis (RQ6)}

\begin{figure}[t]
    \centering
  \begin{subfigure}{.49\linewidth}
  \centering
  \includegraphics[width=0.9\linewidth]{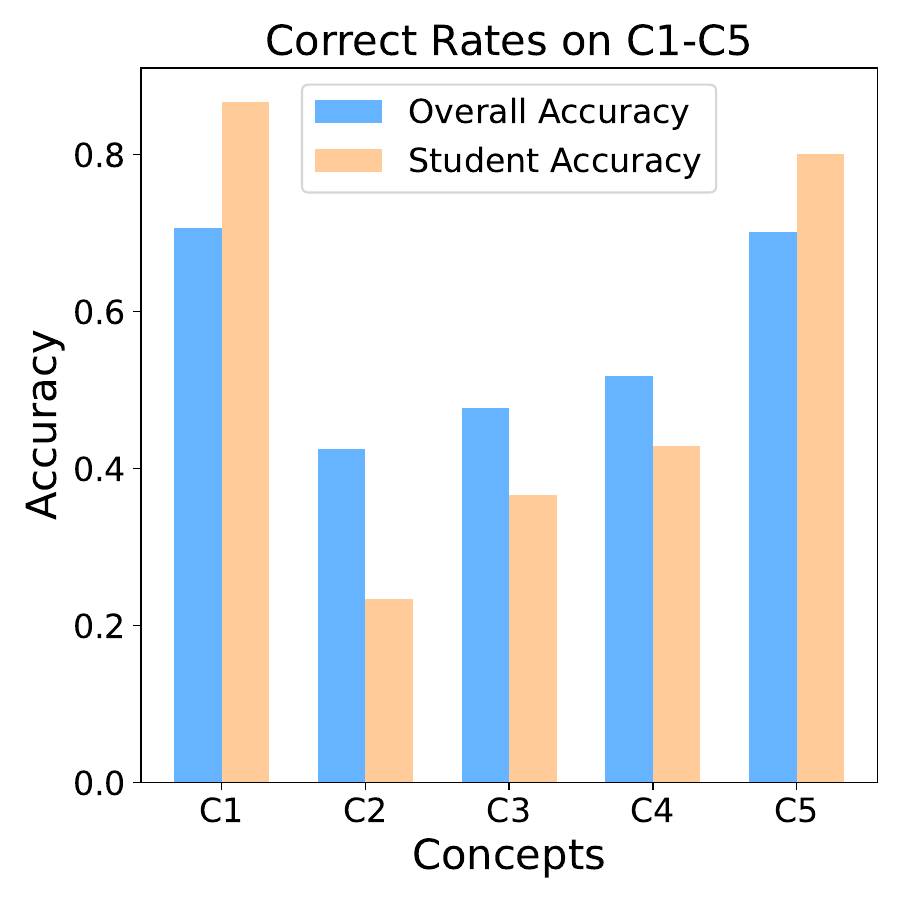}
    \caption{Correct rates on C1-C5}
    \label{fig:case1}
    \end{subfigure}
     \begin{subfigure}{.49\linewidth}
     \centering
  \includegraphics[width=0.9\linewidth]{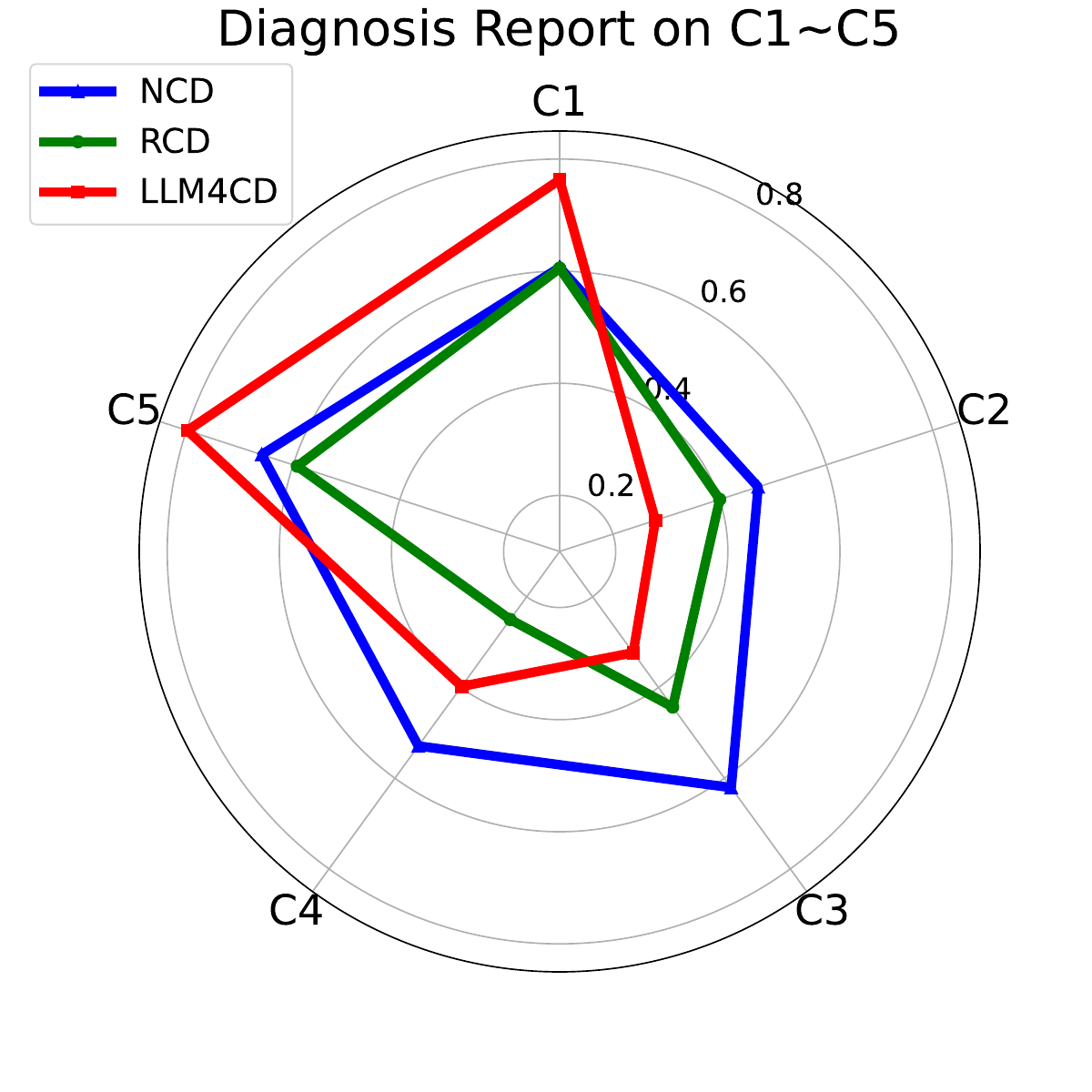}
    \caption{Diagnostic results on C1-C5}
    \label{fig:case2}
    \end{subfigure}
    \vspace{-5pt}
    \caption{An example of a student's diagnostic report.}
    \vspace{-5pt}
\end{figure}
% For a deeper look on the diagnostic ability of LLM4CD, we conduct a case study on a new student unseen in the training set. We select five concepts from the ASSIST09 dataset. The student chosen for this case study was not included in the training data but had a relevant test history. As illustrated in Figure~\ref{fig:case1}, we first analyze the overall correctness rate for these concepts in the dataset and the correctness rate of this student on the same concepts. Next, we generate diagnosis reports for the student using three different models: LLM4CD, NCD, and RCD. As shown in Figure~\ref{fig:case2}, we can see that the diagnosis reports generated by the previous methods tend to be smooth, confined mainly to the 0.4 to 0.6 range, while the diagnosis reports generated by LLM4CD tend to be sharp and discrimitive. The results indicate that NCD and RCD models struggle to accurately capture the student's cognitive state regarding these concepts due to their reliance on student IDs. In contrast, LLM4CD, by constructing text-based models from the student's own test history, provided more precise and discriminative diagnoses that were also more closely aligned with the student's ground truth performance. The results demonstrate that, owing to its ability to model using textual data, LLM4CD offers better interpretable insights into diagnosing student knowledge states, even in a cold-start scenario.

To gain deeper insight into the diagnostic ability of LLM4CD, we conduct a case study on a new student who was not included in the training set. Five concepts from the ASSIST09 dataset are selected for this analysis. Although the selected student was unseen during training, his test history contains relevant interactions that enable the model to derive a meaningful student proficiency representation. As illustrated in Figure~\ref{fig:case1}, we first analyze the overall correctness rate of these concepts within the dataset and compare it to the correctness rate of this specific student. Next, diagnosis reports are generated for the student using three different models: LLM4CD, NCD, and RCD. As shown in Figure~\ref{fig:case2}, the diagnosis reports generated by NCD and RCD are relatively smooth, with predictions confined mostly to the range of 0.4 to 0.6. This indicates that these models, which rely heavily on ID embeddings, struggle to capture the student's true cognitive state. In contrast, the diagnosis reports generated by LLM4CD are sharper and more discriminative, effectively distinguishing the student's proficiency across different concepts. 

These results demonstrate that LLM4CD not only provides more precise and discriminative diagnoses but also aligns more closely with the student's actual performance. By leveraging text-based representations instead of ID-based embeddings, LLM4CD offers interpretable and robust insights into diagnosing student knowledge states, even in cold-start scenarios.

\section{CONCLUSION}
In this paper, we introduce LLM4CD, an advanced CD framework that effectively addresses two critical limitations of traditional ID-based models. First, by leveraging the open-world knowledge of LLMs, we enrich the textual information used in CD tasks, overcoming the constraints of relying solely on ID embeddings. Second, our bi-level encoder structure, which operates at both macro and micro levels, ensures that nuanced semantic features are captured and integrated into the CD task. Our approach is the first to integrate LLMs' open-world knowledge into the CD domain, demonstrating superior performance over current ID-based models and effectively addressing the cold-start problem for new students and exercises. Additionally, we explored how to construct texts within the CD context to maximize the extraction of cognitive information by LLMs. Extensive experimental validation confirmed the effectiveness and robustness of each framework component. As educational contexts continue to generate rich textual data, we will further integrate the advantages of LLMs into CD tasks, aiming to develop even more sophisticated and adaptive CD tools for future ITS.

\bibliographystyle{ACM-Reference-Format}
\bibliography{sample-base}

\end{document}